%% file: PrePrint_Version.tex
\documentclass[preprint,12pt,authoryear]{elsarticle}

\usepackage[utf8]{inputenc}
\usepackage[T1]{fontenc}
\usepackage{graphicx}
\usepackage{booktabs}
\usepackage{amsfonts}
\usepackage{nicefrac}
\usepackage{microtype}
\usepackage{amsmath}
\usepackage{amssymb}
\usepackage{doi}
\usepackage[nolist,nohyperlinks]{acronym}
\usepackage{multirow}
\usepackage{threeparttable}
\usepackage{tabularx}
\usepackage[table]{xcolor}
\usepackage{subcaption}
\usepackage{hyperref}

\input{Acro.tex}
\usepackage{orcidlink}

\journal{Applied Soft Computing}
\date{}
\makeatletter
\def\ps@pprintTitle{}
\makeatother

\begin{document}

\begin{frontmatter}

\title{Forecasting Coccidioidomycosis (Valley Fever) in Arizona: A Graph Neural Network Approach}

\author[asu1]{Ali Sarabi}
\ead{asarabi1@asu.edu}
\author[asu1]{Arash Sarabi}
\ead{sarabi.arash@asu.edu}
\author[asu1]{Hao Yan}
\ead{haoyan@asu.edu}
\author[asu2]{Beckett Sterner}
\ead{beckett.sterner@asu.edu}
\author[asu3]{Petar Jevti\'c}
\ead{petar.jevtic@asu.edu}

\affiliation[asu1]{organization={School of Computing and Augmented Intelligence, Arizona State University},
            city={Tempe},
            postcode={85281}, 
            state={AZ},
            country={USA}}

\affiliation[asu2]{organization={School of Life Sciences, Arizona State University},
            city={Tempe},
            postcode={85281}, 
            state={AZ},
            country={USA}}

\affiliation[asu3]{organization={School of Mathematical and Statistical Sciences, Arizona State University},
            city={Tempe},
            postcode={85281}, 
            state={AZ},
            country={USA}}

\begin{abstract}
Coccidioidomycosis, commonly known as Valley Fever, remains a significant public health concern in endemic regions of the southwestern United States. This study develops the first \ac{GNN} model for forecasting Valley Fever incidence in Arizona. The model integrates surveillance case data with environmental predictors using graph structures, including soil conditions, atmospheric variables, agricultural indicators, and air quality metrics. 
Our approach explores correlation-based relationships among variables influencing disease transmission. The model captures critical delays in disease progression through lagged effects, enhancing its capacity to reflect complex temporal dependencies in disease ecology.
Results demonstrate that the \ac{GNN} architecture effectively models Valley Fever trends and provides insights into key environmental drivers of disease incidence. These findings can inform early warning systems and guide resource allocation for disease prevention efforts in high-risk areas.\footnote{Code and data are available at \url{https://github.com/a-sarabi/Valley-Fever}}
\end{abstract}


\end{frontmatter}

\section{Introduction}
Coccidioidomycosis, commonly known as Valley Fever, is a fungal respiratory disease endemic to the arid regions of the southwestern United States, particularly Arizona \citep{maricopa_valley_fever}. Caused by the soil-dwelling fungi Coccidioides spp., the disease poses a significant public health challenge due to its increasing incidence and expanding geographic range within the state \citep{hector2011public}. Over the past two decades, Arizona has witnessed a notable rise in reported cases \citep{cdc_valley_fever}, underscoring the need for a deeper understanding of the factors driving disease transmission and spread.

Many infectious diseases exhibit seasonality and spatial patterns \citep{CID,comrie2005climate} that are influenced by environmental and climatic factors \citep{del2016habitat}. Understanding these patterns is crucial for identifying the mechanisms underlying disease dynamics and for developing effective public health interventions. For coccidioidomycosis, the relationship between environmental conditions and pathogen dynamics remains complex and debated. The traditional hypothesis views Coccidioides as primarily soil-dwelling fungi, where wet conditions promote fungal growth in soil and dry, windy conditions facilitate dispersal of infectious arthroconidia through dust particles \citep{head2022effects}. However, an alternative endozoan hypothesis proposes that Coccidioides species primarily live as endozoans in small mammals, residing as spherules in host granulomas and sporulating when the host dies \citep{taylor2019endozoan}. 
Under this hypothesis, environmental factors may exert indirect effects on disease incidence through complex ecological pathways. Climate and soil conditions affect plant growth, which in turn influences small mammal populations that serve as the primary reservoir for the fungi.

Previous studies have highlighted the association between environmental factors and coccidioidomycosis incidence. However, the complex interplay of temporal variables and their lagged effects on disease transmission remains less understood. Traditional statistical models often fall short in capturing these intricate dependencies and typically rely on manual feature selection, necessitating the application of advanced computational methods \citep{madakkatel2021combining}.

In this study, we employ a \ac{GNN} model to forecast Valley Fever incidence in Arizona from 2006 to 2024. GNNs excel at modeling relational data by representing complex systems as interconnected nodes and edges, making them ideally suited for capturing the temporal dependencies inherent in infectious disease dynamics. Our approach integrates comprehensive surveillance data with an extensive array of environmental predictors.
A key feature of our approach is the incorporation of lagged effects to account for delays in disease onset, diagnosis, and reporting. This enhances the model's ability to reflect real-world temporal dependencies and provides more accurate forecasts.

By advancing predictive modeling techniques and deepening our comprehension of the environmental drivers of coccidioidomycosis, this work contributes to efforts aimed at mitigating the impact of Valley Fever in Arizona. The findings have important implications for public health preparedness, enabling stakeholders to anticipate disease surges and implement timely preventive measures in the face of changing environmental conditions.

\section{Background}

Coccidioidomycosis, also known as Valley fever, poses major public health challenges in endemic regions due to its complex environmental dependencies and unpredictable transmission dynamics. This section provides an overview of the disease’s ecological background, the environmental and climatic factors driving its spread, and recent advances in computational modeling approaches used to improve forecasting.

\subsection{Coccidioidomycosis Ecology and Modeling Challenges}
Coccidioidomycosis represents a complex infectious disease system where environmental, temporal, and spatial factors interact in ways that traditional epidemiological models struggle to capture effectively. The disease's unique ecology, characterized by a soil-dwelling pathogen with specific environmental requirements for growth and dispersal, creates modeling challenges that extend beyond conventional time series forecasting approaches.

The pathogen Coccidioides exists in two primary species, \textit{C. immitis} and \textit{C. posadasii}, which show distinct geographic distributions within endemic regions \citep{dobos2021using}. These fungi undergo a complex lifecycle where environmental conditions dictate both the vegetative growth phase in soil and the formation of infectious arthroconidia. Research has established that the pathogen thrives in alkaline soils with specific mineral compositions, particularly those found in the Lower Sonoran life zone \citep{coopersmith2017relating}. The temporal dynamics of fungal growth and spore release create inherent delays between environmental exposure and clinical manifestation. Incubation periods range from one to three weeks, followed by additional delays in diagnosis and reporting \citep{benedict2018enhanced}.

Environmental modeling of coccidioidomycosis has evolved from simple correlation studies to more complex analyses incorporating multiple predictor variables. Early research primarily focused on meteorological factors, particularly precipitation and temperature patterns \citep{comrie2005climate}. Tamerius and Comrie demonstrated that seasonal precipitation patterns could predict annual coccidioidomycosis incidence in Arizona, establishing the foundation for environmental forecasting approaches \citep{tamerius2011coccidioidomycosis}. Subsequent studies expanded this framework to include soil moisture dynamics, drought indices, and air quality metrics, recognizing that dust events serve as primary vectors for spore dispersal \citep{lauer2020valley}.

The spatial heterogeneity of coccidioidomycosis presents particular challenges for traditional modeling approaches. Disease incidence varies significantly across geographic scales, from county-level variations to fine-scale neighborhood differences within endemic areas \citep{brown2017spatial}. This spatial variation reflects complex interactions between soil characteristics, land use patterns, population density, and microclimatic conditions. Previous attempts to model these spatial patterns using conventional regression techniques have shown limited success in capturing the full complexity of these relationships \citep{weaver2020environmental}.

Temporal modeling of coccidioidomycosis has historically relied on autoregressive integrated moving average (ARIMA) models and related time series approaches. While these methods can capture basic seasonal patterns and trends, they face fundamental limitations when applied to coccidioidomycosis data. The disease exhibits irregular seasonality that varies between years and locations, driven by complex interactions between multiple environmental variables operating at different temporal scales \citep{head2022effects}. Traditional time series models assume linear relationships and stationary data properties that may not hold for infectious disease systems influenced by climate variability and extreme weather events.

These limitations in conventional approaches have led researchers to explore machine learning applications in infectious disease epidemiology, particularly for diseases with complex environmental dependencies. LSTM networks have shown promise for epidemic forecasting, as demonstrated in recent COVID-19 and influenza prediction studies \citep{ye2025integrating}. However, these approaches typically treat environmental variables as independent inputs, failing to capture the complex interdependencies that characterize environmental systems. This limitation is particularly relevant for environmentally-driven diseases like coccidioidomycosis, where multiple environmental factors interact through complex correlation structures and feedback mechanisms.

Graph Neural Networks represent a paradigm shift in multivariate modeling by explicitly incorporating relational information between system components. GNNs model systems as interconnected networks where nodes can represent either spatial entities or system variables, and edges encode relationships between them \citep{wu2020comprehensive}. While traditionally applied to spatial networks, this architecture is equally well-suited for modeling variable interactions in complex environmental systems, where multiple factors influence outcomes through intricate correlation structures and interdependencies.

\subsection{Environmental Factors and Climate Change Impacts}

Recent research has revealed expanding geographic distribution of coccidioidomycosis driven by climate change, with Arizona reporting a record 14,680 cases in 2024 compared to 9,148 in 2023 \citep{vax_before_travel_2025}. Climate modeling studies project that the Coccidioides endemic region will expand from the southwestern United States to much of the western United States by 2100, with specific climate thresholds identified as annual temperature greater than 10.7°C and annual precipitation less than 600mm annually. A 50\% increase in cases is predicted by 2100 under high-warming scenarios, with new endemic states projected to include Montana, Nebraska, North Dakota, South Dakota, and Wyoming \citep{gorris2019expansion}.

The correlation between drought-wet cycles and increased incidence has been quantified through recent studies, with successful prediction models based on temperature and precipitation data. Foundational research by \citet{tamerius2011coccidioidomycosis} demonstrated strong positive correlations between October-December precipitation and exposure rates in both Maricopa and Pima counties, with models explaining up to 69\% of disease variance. Recent analysis by \citet{head2022effects} quantified drought effects across the western United States, estimating thousands of excess cases following major drought periods and demonstrating that winter precipitation effects are amplified when preceded by dry conditions. Research has identified that Arizona's bimodal precipitation pattern creates distinct seasonality with peaks in June-July and October-November \citep{comrie2005climate}, while surface soil moisture levels are critical for soil-borne fungal pathogen survival and spore production \citep{coopersmith2017relating}.

Current environmental modeling for Valley fever forecasting integrates additional atmospheric and ecological variables that complement traditional temperature and precipitation measurements. Wind speed and direction patterns are increasingly recognized as important factors for spore dispersal, though research shows complex relationships with disease transmission that vary by geographic location \citep{comrie2021no}. El Niño Southern Oscillation patterns demonstrate significant correlations with soil moisture conditions and subsequent disease incidence through multi-year climate cycles \citep{tobin2022coccidioidomycosis}. Air quality metrics have emerged as important environmental indicators for Valley fever variability, with atmospheric conditions influencing disease transmission patterns in Arizona's desert environment. Dust storm patterns present complex relationships with disease transmission, though recent research reveals conflicting findings regarding direct dust storm-disease associations \citep{comrie2021no}. These environmental variables enhance understanding of the complex ecological relationships governing Coccidioides transmission in Arizona's desert regions.

Environmental monitoring for fungal disease prediction has advanced through the integration of satellite remote sensing and climate data. While hyperspectral imaging shows promise for pathogen detection in agricultural applications \citep{ferreira2024hyperspectral}, with high classification accuracies for various fungal pathogens, specific applications to Coccidioides environmental surveillance in Arizona's arid environment remain an area for future research. These advances provide critical insights into how climate variables and soil conditions influence Valley fever transmission dynamics in Arizona's desert regions.

\subsection{Deep Learning and Multi-Modal Approaches}

Machine learning has transformed disease forecasting, with Convolutional Neural Networks, Random Forest, and Support Vector Machines demonstrating substantial improvements over conventional methods \citep{ganjalizadeh2023machine}. Transformer architectures with self-attention mechanisms have advanced disease prediction by better capturing long-term dependencies in temporal data \citep{yang2023transformehr, lv2024evaluating}. Hybrid frameworks combining convolutional and recurrent components have excelled in epidemiological applications \citep{zain2021covid}, leveraging spatial feature extraction alongside temporal pattern recognition to create comprehensive progression models enhanced by optimization algorithms for hyperparameter tuning. Time-aware architectures achieve exceptional accuracy with irregular time series data \citep{sun2021predicting}, while ensemble strategies combining different deep learning frameworks through voting and stacking methods show statistically significant improvements. Bayesian Model Averaging provides uncertainty quantification that accounts for model variability.

Spatial-temporal modeling has evolved to capture both geographic transmission patterns and temporal dynamics simultaneously. Spectral approaches model dependencies using Fourier transforms \citep{tomy2022estimating}, while dynamic graph networks incorporate mobility patterns for enhanced forecasting performance. Multi-scale modeling through learning-to-cluster algorithms captures local and global signals, outperforming exclusively spatial or temporal approaches.

For coccidioidomycosis, variable-based graph approaches offer distinct advantages by explicitly modeling interdependencies between environmental variables while incorporating heterogeneous data types within unified frameworks. These methods can learn adaptive relationships rather than relying on predefined structures, potentially discovering novel environmental interaction patterns. The disease exhibits complex lagged effects, where environmental conditions in preceding months influence current incidence. Precipitation effects are typically observed with lags of several months, while temperature and wind patterns show more immediate but less consistent associations \citep{weaver2018investigating}. Graph architectures can learn optimal lag structures dynamically during training, unlike traditional statistical approaches using fixed temporal offsets.

However, environmental data integration poses challenges related to quality, resolution, and coverage. Meteorological data may inadequately represent conditions across large areas with complex topography \citep{dlamini2019review}, while satellite-derived products offer improved spatial coverage but may lack temporal resolution needed for forecasting. The complexity of coccidioidomycosis ecology necessitates approaches handling non-linear relationships, temporal dependencies, and spatial correlations simultaneously, with model interpretability being crucial for public health applications.

This study addresses these challenges by developing a graph neural network framework specifically designed for coccidioidomycosis forecasting that incorporates comprehensive environmental data, accounts for inter-variable dependencies, and models complex temporal relationships including lagged effects.

\subsection{Graph Neural Networks in Infectious Disease Modeling}
Recent advances in \ac{GNN} applications have demonstrated their effectiveness in modeling disease transmission patterns. Traditional spatial applications excel at learning complex spatiotemporal patterns by aggregating information from neighboring geographic nodes while maintaining the ability to capture temporal dependencies through recurrent architectures.

The period from 2020 to 2024 has witnessed significant developments in \ac{GNN} applications to infectious disease modeling, driven largely by the COVID-19 pandemic's urgent demands for better forecasting capabilities. \ac{GCN}s have demonstrated substantial improvements in disease prediction tasks when compared to traditional epidemiological models, with successful applications in COVID-19 forecasting, dengue prediction, and influenza surveillance \citep{kapoor2020examining,song2023covid}. More advanced architectures, including \ac{GAT}s and the Graph Attention-based Spatial Temporal (GAST) model, employ attention mechanisms to provide refined understanding of dynamics through analysis frameworks.

Recent innovations have focused on hybrid architectures that combine GNNs with recurrent neural networks through Recurrent Message Passing (RMP) modules. These frameworks utilize Gated Recurrent Units for temporal modeling and Graph Attention Networks for relational interactions, showing strong performance for COVID-19 spread prediction \citep{liu2024review}. Attention-based Multiresolution Graph Neural Networks (ATMGNN) have further advanced the field by learning to combine relational graph information with temporal timeseries data using clustering algorithms to capture multiscale structures, enabling modeling of long-range dependencies.

Current research has developed sophisticated approaches for constructing meaningful graphs from diverse data sources. Variable interaction networks can be constructed based on correlation structures, mutual information, or causal relationships, enabling the modeling of complex environmental systems. Graph construction methodologies now include adaptive edge weighting schemes that allow relationships to evolve during training, and hierarchical graph structures that capture interactions at multiple scales. These variable-centric approaches have demonstrated significant improvements over traditional feature selection methods through enhanced modeling of feature interdependencies \citep{liu2024review}.

\section{Methodology}
\begin{figure}[htbp]
    \centering
    \includegraphics[width=1\linewidth]{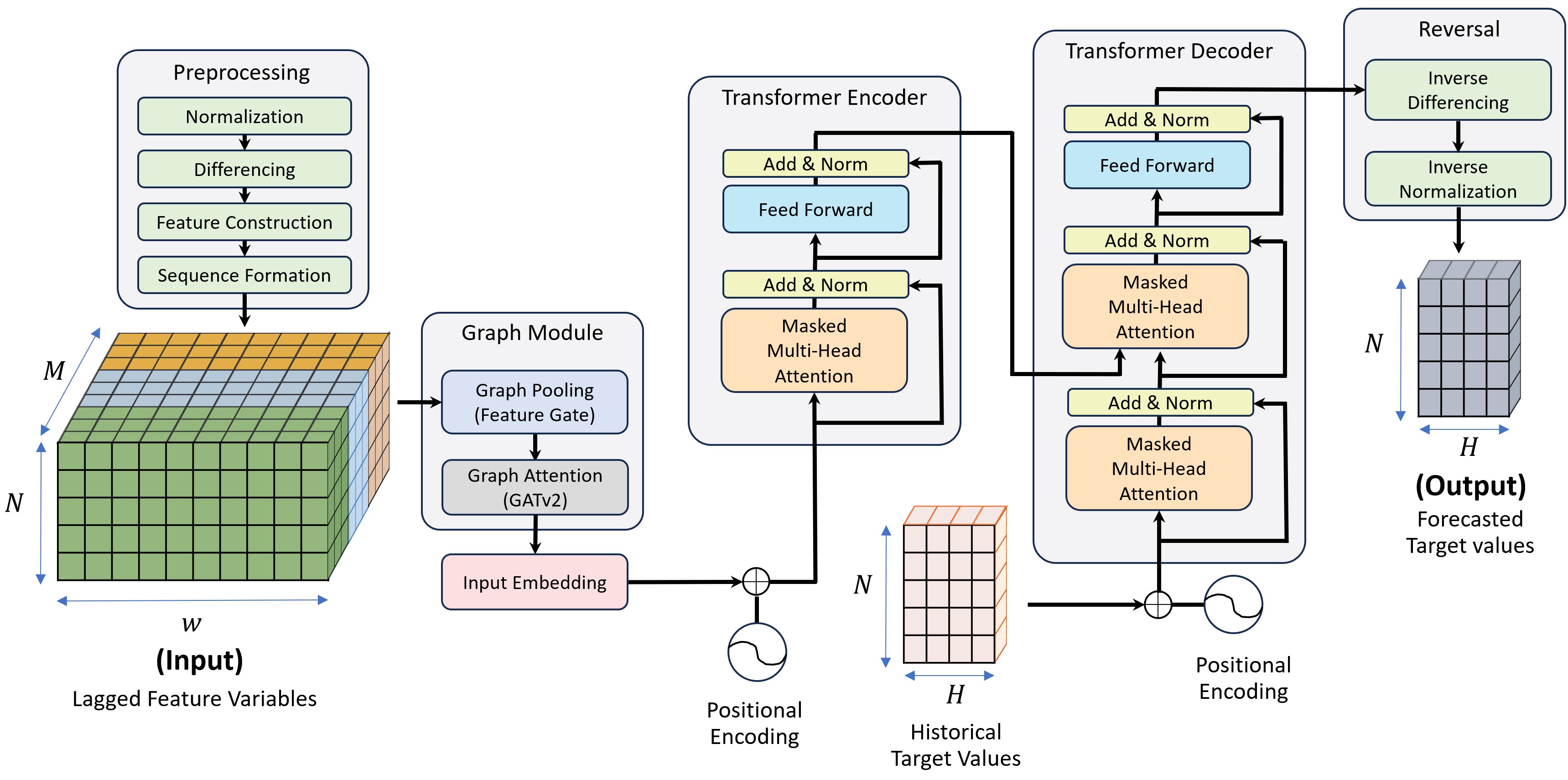}
    \caption{Overview of the proposed forecasting model architecture. The pipeline integrates an Improved Feature Gate for adaptive feature selection, Graph Attention Networks (GATv2) for spatial dependency modeling, and a Transformer-based encoder-decoder for temporal sequence prediction.}
    \label{fig:model_architecture}
\end{figure}

In this section, we detail our forecasting methodology, built on a hybrid architecture that integrates a Graph Neural Network (GNN) with a Transformer to capture complex spatiotemporal dependencies. Our approach is organized into two main components. In Section~\ref{ssec:graph_module}, we describe the graph module, which models the relationships between environmental variables. This includes our method for graph construction, feature selection via a graph pooling layer, and learning variable interactions using Graph Attention Networks. In Section~\ref{ssec:transformer_architecture}, we detail the Transformer-based encoder-decoder architecture that processes these feature representations to model temporal dynamics and generate multi-step forecasts. Figure~\ref{fig:model_architecture} provides an overview of the proposed forecasting model architecture.

\subsection{Graph Module}\label{ssec:graph_module} 

\begin{figure}[h]
\centering
\includegraphics[width=1\textwidth]{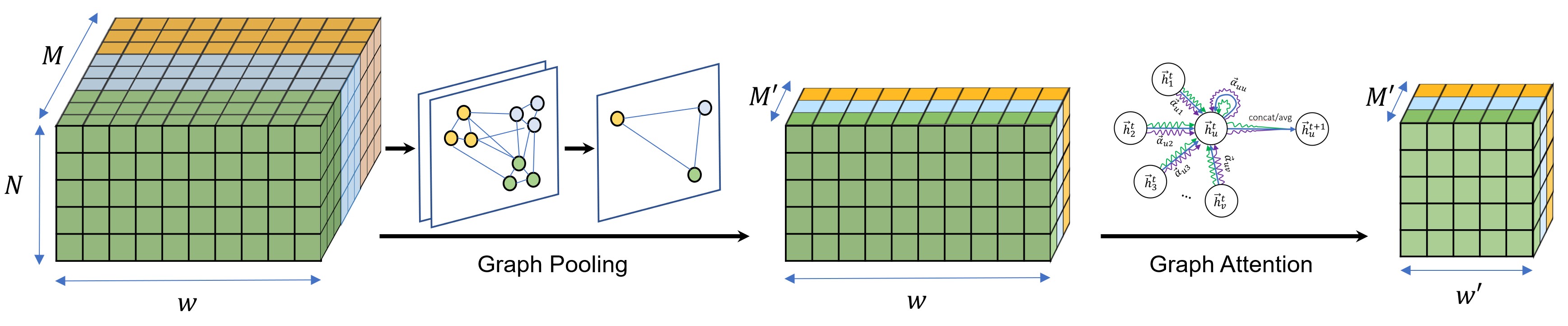}
\caption{Graph module. The graph is first constructed using feature correlations (left), followed by graph pooling layers to retain the top 10\% of informative features (middle), and finally processed with graph attention layers to learn attention-weighted representations (right).}
\label{fig:graph_module}
\end{figure}

The graph construction component forms the foundation of our modeling approach.  
We represent feature relationships through a weighted graph structure where nodes correspond to individual variables—both original and lagged—and edges are weighted by Pearson correlation coefficients.  
Correlations whose magnitude falls below a threshold of 0.05 are set to zero, reducing noise while preserving meaningful connections.  
This graph allows the model to capture complex interdependencies and delayed effects among environmental factors.  
Figure~\ref{fig:graph_module} provides a visual overview of the full pipeline, including graph construction, graph pooling, and graph attention layers.

Formally, let \(X=\{x_{m,t}\}\in\mathbb{R}^{M\times T}\) be the multivariate time series; the Pearson coefficient between series \(u\) and \(v\) is
\[
\rho_{uv}
= \frac{\sum_{t=1}^T (x_{u,t}-\bar{x}_u)(x_{v,t}-\bar{x}_v)}
       {\sqrt{\sum_{t=1}^T(x_{u,t}-\bar{x}_u)^2}\;
        \sqrt{\sum_{t=1}^T(x_{v,t}-\bar{x}_v)^2}},
\]
and the adjacency weights are
\[
a_{uv}= \begin{cases}
\lvert\rho_{uv}\rvert,&\lvert\rho_{uv}\rvert\ge 0.05,\\
0,&\text{otherwise}.
\end{cases}
\]

\paragraph{Graph Pooling (Feature Gate)}
The Feature Gate is implemented as a graph-pooling mechanism that dynamically identifies and retains the most informative features during training. By selecting only the top 10\% of features based on their learned importance, this pooling operation significantly reduces model complexity while preserving predictive power. This approach allows the model to focus computational resources on the most relevant environmental indicators while preserving the relational structure among them for subsequent attention operations.

Formally, each node $i$ is assigned a trainable logit, $w_i$. Applying a sigmoid function yields a continuous gate $g_i = \sigma(w_i) \in (0, 1)$. At every forward pass, we keep the $k = \lceil pM \rceil$ largest gates (with $p = 0.10$ in our experiments) and set the remaining ones to zero, producing the masked vector $\tilde{g}$. The Feature Gate then multiplies the node-feature tensor $H$ by $\tilde{g}$ (where $\tilde{g}_i = g_i$ for the selected nodes and $\tilde{g}_i = 0$ otherwise). Given a batch of node-feature tensors $H \in \mathbb{R}^{B \times M \times d}$, the Feature Gate operates as

\begin{equation}
\Phi_{\text{gate}}(H) = H \;\odot\; \tilde{g},
\end{equation}
where $\odot$ denotes element-wise multiplication, broadcast across the batch and feature dimensions. In practice, this operation prunes about 90\% of the variables while softly re-weighting the remaining ones, adding negligible computational overhead.

\paragraph{Graph Attention Networks (GATv2)}

After selecting relevant features, we apply GATv2 layers to learn representations of the environmental variables, each represented as a node in the graph. This formulation allows the model to attend over the neighborhood of each variable, dynamically weighting its connections to others based on their contribution to the prediction task. Such adaptive weighting is particularly valuable for capturing how multiple environmental factors interact in disease transmission dynamics.

In the standard GAT architecture, attention coefficients $\alpha_{ij}$ are defined as:

\begin{equation}
\alpha_{ij}
= \frac{\exp\!\bigl(\operatorname{LeakyReLU}(\mathbf a^{\!\top}[\,W h_i \,\|\, W h_j])\bigr)}
       {\sum_{k\in\mathcal N(i)}\exp\!\bigl(\operatorname{LeakyReLU}(\mathbf a^{\!\top}[\,W h_i \,\|\, W h_k])\bigr)},
\end{equation}
where $h_i$ denotes the feature vector associated with node $i$ (i.e., an environmental variable), $W$ is a shared linear transformation, and $\mathbf{a}$ is a learnable attention vector. The neighbourhood $\mathcal{N}(i)$ includes all variables correlated with $i$, and attention scores are normalised via softmax.

GATv2 generalises this formulation by decoupling the feature transformations applied to the source and target nodes, enabling greater flexibility in modelling directional and asymmetric dependencies:

\begin{equation}
\alpha_{ij}
= \frac{\exp\!\bigl(\mathbf a^{\!\top}\operatorname{LeakyReLU}(\mathbf W_1 h_i+\mathbf W_2 h_j)\bigr)}
       {\sum_{k\in\mathcal N(i)}\exp\!\bigl(\mathbf a^{\!\top}\operatorname{LeakyReLU}(\mathbf W_1 h_i+\mathbf W_2 h_k)\bigr)}.
\end{equation}

Here, $\mathbf{W}_1$ and $\mathbf{W}_2$ are separate learnable projection matrices for the source and target nodes, respectively. This decoupled attention mechanism allows the model to capture subtle variations in how each environmental factor both influences and is influenced by others, a crucial property when modelling temporally lagged processes such as disease transmission.

Once attention coefficients are computed, node representations are updated using a weighted aggregation over neighbours. For multi-head attention with $H$ heads, this is given by:

\begin{equation}
h_i^{(l+1)} = \mathbin\Vert_{k=1}^{H} \sigma\left( \sum_{j \in \mathcal N(i)} \alpha_{ij}^{(k)}\, W^{(k)} h_j^{(l)} \right),
\end{equation}
where $\sigma$ is a non-linear activation function (e.g., ReLU), and $\Vert$ denotes concatenation of the outputs from each attention head. Each head has its own weight matrix $W^{(k)}$ and attention scores $\alpha_{ij}^{(k)}$, allowing the model to capture diverse interaction patterns between variables.

\subsection{Transformer Architecture}\label{ssec:transformer_architecture}

To model long-range temporal dependencies, we implement a Transformer-based encoder-decoder architecture that operates on temporally ordered embeddings of environmental features. These embeddings are derived from the graph module after spatial processing via GATv2 layers.

Let the temporal input sequence for forecasting at time $t$ be $\mathbf{Z}_{t} = [\mathbf{z}_{t-w}, \ldots, \mathbf{z}_{t-1}] \in \mathbb{R}^{w \times d}$, where each $\mathbf{z}_{\tau} \in \mathbb{R}^{d}$ represents the vector embedding at time step $\tau$, output from the GAT-based spatial encoder. Here, $w$ is the sequence length and $d$ is the dimensionality of the latent feature representation. To preserve temporal information, we add positional encodings $\mathbf{P} \in \mathbb{R}^{w \times d}$ to each time step, yielding

\begin{equation}
\tilde{\mathbf{Z}}_{t} = \mathbf{Z}_{t} + \mathbf{P}.
\end{equation}

The Transformer encoder applies multi-head self-attention to $\tilde{\mathbf{Z}}_{t}$, computing pairwise interactions across all positions in the sequence. For each attention head, the computation is defined as:

\begin{equation}
\text{Attention}(\mathbf{Q}, \mathbf{K}, \mathbf{V}) = \text{softmax}\left( \frac{\mathbf{Q} \mathbf{K}^{\top}}{\sqrt{d_k}} \right)\mathbf{V},
\end{equation}
where $\mathbf{Q} = \tilde{\mathbf{Z}}_{t}W_Q$, $\mathbf{K} = \tilde{\mathbf{Z}}_{t}W_K$, and $\mathbf{V} = \tilde{\mathbf{Z}}_{t}W_V$ are query, key, and value projections with learnable matrices $W_Q, W_K, W_V \in \mathbb{R}^{d \times d_k}$. Outputs from all attention heads are concatenated and passed through a feed-forward network with residual connections and layer normalization, resulting in encoder output $\mathbf{Z}_{\text{enc}} \in \mathbb{R}^{w \times d}$.

The decoder receives the previously observed or predicted target values $\mathbf{y}_{t-w:t-1}$, embedded and augmented with positional encodings. It applies masked self-attention to preserve causal structure and cross-attention to $\mathbf{Z}_{\text{enc}}$ to generate predictions over a forecast horizon $H$:

\begin{align}
\mathbf{Z}_{\text{dec}} &= \text{DecoderSelfAttention}(\mathbf{Y}_{\text{in}} + \mathbf{P}), \\
\hat{\mathbf{y}} &= \text{DecoderCrossAttention}(\mathbf{Z}_{\text{dec}}, \mathbf{Z}_{\text{enc}}).
\end{align}

Because the model is trained on differenced target values, a reverse differencing layer is applied to recover the final predictions:

\begin{equation}
\hat{y}_{t+h} = \hat{\Delta}_{t+h} + y_{t+h-1}, \quad \text{for } h = 0, 1, \ldots, H-1.
\end{equation}

This Transformer module is trained jointly with the graph components using backpropagation and a \ac{MSE} loss computed over multi-step predictions. It enables the model to capture fine-grained temporal dynamics in disease progression conditioned on spatially encoded environmental features.

\section{Experiments and Results} \label{section.experiments}
In this section, we detail the experimental setup and present the results that validate the effectiveness of our proposed framework. We begin by describing the data collection methods (see Section \ref{ssec:data_collection}) and preprocessing steps (see Section \ref{ssec:preprocessing}) undertaken to harmonize diverse sources, including disease surveillance records, environmental measurements, and air quality metrics. These preparations ensure that our time series inputs comprehensively capture the temporal and spatial variability inherent in Valley Fever incidence.

Following data preparation, we outline the training protocols (see Section \ref{ssec:training}) and evaluation strategies used to benchmark model performance. Experiments were conducted over multiple forecast horizons to assess not only predictive accuracy but also the robustness of the graph-based feature selection (see Section \ref{ssec:feature_importance}) and Transformer-based temporal encoding approaches. The results showcase the model's ability to capture both intricate spatiotemporal interdependencies and delayed environmental influences on disease transmission.

\subsection{Data Collection and Sources}\label{ssec:data_collection}

Our dataset comprises Valley Fever surveillance data from Maricopa County, Arizona, along with environmental predictors, including soil conditions (20" and 4" soil temperature measurements), atmospheric variables (air temperature, dewpoint, actual vapor pressure, relative humidity, wind speed and direction, precipitation, solar radiation, and vapor pressure deficit), agricultural indicators (reference evapotranspiration and heat units), and air quality metrics (AQI arithmetic mean). All data were aggregated at weekly intervals following the MMWR epidemiological calendar from 2006 to 2024.

\paragraph{Valley Fever Surveillance Data}
Valley fever incidence data for Maricopa County were obtained from the Arizona Department of Health Services\footnote{Arizona Department of Health Services: \url{www.azdhs.gov} } in weekly MMWR (Morbidity and Mortality Weekly Report) format. The MMWR week system is the standardized epidemiological calendar used by the National Notifiable Diseases Surveillance System (NNDSS), where weeks run from Sunday through Saturday and are numbered 1-52 (occasionally 53). Week \#1 of an MMWR year is defined as the first week of the year that contains at least four days in the calendar year. This standardized temporal framework ensures consistent disease reporting and enables accurate temporal analysis of Valley Fever patterns.

\paragraph{Environmental and Weather Data}
Daily weather data were acquired from the AZMET (Arizona Meteorological Network) Weather Data portal\footnote{AZMET Weather Data: \url{https://cales.arizona.edu/AZMET/az-data.htm} }, which provides comprehensive meteorological measurements across Arizona. The daily measurements were then converted and aggregated to weekly MMWR format using custom Python code to align with the epidemiological calendar and ensure temporal consistency with the Valley Fever surveillance data.

\paragraph{Air Quality Data}
Daily PM10 concentrations were obtained from the U.S. Environmental Protection Agency's Air Quality System database\footnote{EPA Air Quality Data: \url{https://www.epa.gov/outdoor-air-quality-data/download-daily-data}}. The daily PM10 measurements were similarly aggregated to weekly MMWR format to maintain temporal alignment across all data sources.

\subsection{Data Preprocessing and Feature Engineering}\label{ssec:preprocessing}

The raw time series data undergoes a multi-step preprocessing pipeline to prepare it for the forecasting model. This includes normalization, differencing, feature construction, and sequence generation.

\paragraph{Normalization and Differencing}
To standardize the scales of different environmental variables, all features are normalized to a [0, 1] range using min-max scaling. This prevents variables with larger magnitudes from disproportionately influencing model training. Furthermore, to ensure the time series is stationary, we apply first-order differencing to the target variable (Valley Fever cases). This step removes underlying trends and stabilizes the statistical properties of the series, helping the model focus on predicting weekly changes. The original case counts are recovered in a post-processing step using reverse differencing.

\paragraph{Lagged Feature and Sequence Construction}
To capture the delayed effects of environmental conditions on Valley Fever incidence, we generate lagged features for all predictors up to a maximum of 6 weeks. This lag window is chosen to account for the disease's incubation period and the time it takes for environmental changes to impact spore dispersal. Following feature creation, we use a sliding window approach to create input-output pairs for training. Each input sample consists of a sequence of length $w$ containing all features from time steps $t-w$ to $t-1$. The corresponding output sample is the sequence of target values from time $t$ to $t+H-1$, where $H$ is the forecast horizon.

\subsection{Training Procedure}\label{ssec:training}

We evaluated the model using a rolling window approach with forecast horizons of 2, 4, 8, and 16 weeks. The training data included all weeks up to week 850, while testing was conducted on weeks 900 to 991, reflecting a realistic forecasting scenario where future data remains unavailable during training.

Model training employed the Adam optimizer with a learning rate of 1e-4 and Mean Squared Error loss function, with training limited to a maximum of 100 epochs. The sequence length was configured to 3 times the forecast horizon, while both the transformer embedding dimension and feed-forward dimension were set to 256. The architecture utilized 8 attention heads with a dropout rate of 0.05 and a feature selection rate of 10\%. Early stopping with a patience of 20 epochs was implemented to prevent overfitting and ensure optimal generalization.

\subsection{Feature Importance Analysis}\label{ssec:feature_importance}
To assess the reliability of our feature selection mechanism performed by our graph pooling layer, we conducted a stability analysis using 100 different random seeds for each forecast horizon. This evaluation provides insights into which environmental variables consistently contribute to Valley Fever forecasting across different time scales. The analysis was performed separately for forecast horizons of 2, 4, 8, and 16 weeks, allowing us to examine how feature importance varies with prediction length.

The graph pooling layer identified the top 10\% most important features in each case, resulting in a 90\% reduction in input dimensionality.
This substantial reduction improves computational efficiency while preserving predictive performance. Table~\ref{tab:top_features_all} presents all features that appear in the top 10\% (19 features) for any forecast horizon. Soil temperature measurements, particularly 20-inch soil temperature maximum values at the current time, emerged as the most reliable predictor, consistently ranking first across all horizons. Relative humidity variables and air quality indicators, such as daily mean PM\textsubscript{10} concentration, also demonstrated high stability and importance. The numbers in parentheses indicate the temporal offset in weeks, where negative values represent past observations and zero denotes the current time.

\begin{table}[h]
\centering
\footnotesize
\caption{All top 10\% features across forecast horizons. Numbers indicate rank (1-19) within each horizon; empty cells indicate the feature was not in the top 10\%.}
\label{tab:top_features_all}
\begin{tabular}{lcccc}
\hline
\textbf{Feature Name (Lag)} & \textbf{2 weeks} & \textbf{4 weeks} & \textbf{8 weeks} & \textbf{16 weeks} \\
\hline
20\textquotedbl{} Soil Temp - Max (0) & 1 & 1 & 1 & 1 \\
RH - Min (-1) & 2 & 2 & 2 & 3 \\
Daily Mean PM\textsubscript{10} Concentration (-6) & 3 & 3 & 3 & 2 \\
RH - Max (-1) & 4 & 4 & 4 & 6 \\
Solar Rad. - Total (-2) & 7 & 5 & 5 & 4 \\
Wind Vector Direction (-1) & 5 & 7 & 6 & 5 \\
Wind Vector Direction (0) & 8 & 8 & 8 & 7 \\
MARICOPA (-1) & 6 & 13 & 7 & 8 \\
Actual Vapor Pressure - Daily Mean (-2) & 10 & 6 & 9 & 13 \\
Reference ET (0) & 9 & 10 & 12 & 9 \\
20\textquotedbl{} Soil Temp - Mean (0) & 12 & 15 & 11 & 10 \\
RH - Max (-6) & 17 & 11 & 10 & 14 \\
Wind Speed - Mean (-5) & 11 & 18 & 13 & 12 \\
4\textquotedbl{} Soil Temp - Max (-5) & 15 & 16 & 17 & 15 \\
Dewpoint - Daily Mean (-6) & 16 & 12 & 16 & 19 \\
Dewpoint - Daily Mean (-5) & 13 & 9 & 14 & - \\
20\textquotedbl{} Soil Temp - Max (-2) & - & 14 & 15 & 16 \\
4\textquotedbl{} Soil Temp - Max (-4) & - & 17 & 19 & 17 \\
Wind Direction Std Dev (-1) & - & - & 18 & 11 \\
Wind Vector Direction (-3) & 18 & - & - & 18 \\
RH - Mean (-1) & 14 & - & - & - \\
Dewpoint - Daily Mean (-1) & 19 & - & - & - \\
RH - Mean (-5) & - & 19 & - & - \\
\hline
\end{tabular}
\end{table}

Across all horizons, 15 features consistently appear in the top 10\% across all four forecast horizons, demonstrating remarkable stability in feature selection. An additional 3 features appear in three horizons, while 5 features are horizon-specific. The majority of selected features fall into the recent lag category ($\leq$3 weeks), including variables from Precipitation/Humidity, Soil, Wind/Temperature/Radiation, Surveillance, and Agricultural domains. However, long-term predictors ($>$3 weeks) also contribute significantly, particularly air quality metrics and certain soil temperature measurements. The temporal analysis reveals that both short-term and long-term environmental signals are essential for accurate Valley Fever forecasting, with their relative importance remaining stable across different prediction horizons.

Notably, soil temperature and humidity variables consistently appear in both recent and long-term categories across all horizons. This indicates that these domains may influence Valley Fever risk through both immediate mechanisms (e.g., spore viability) and longer-term environmental processes (e.g., soil drying cycles that affect spore release and dispersal).

\begin{figure}[htb]
\centering
\includegraphics[width=1\textwidth]{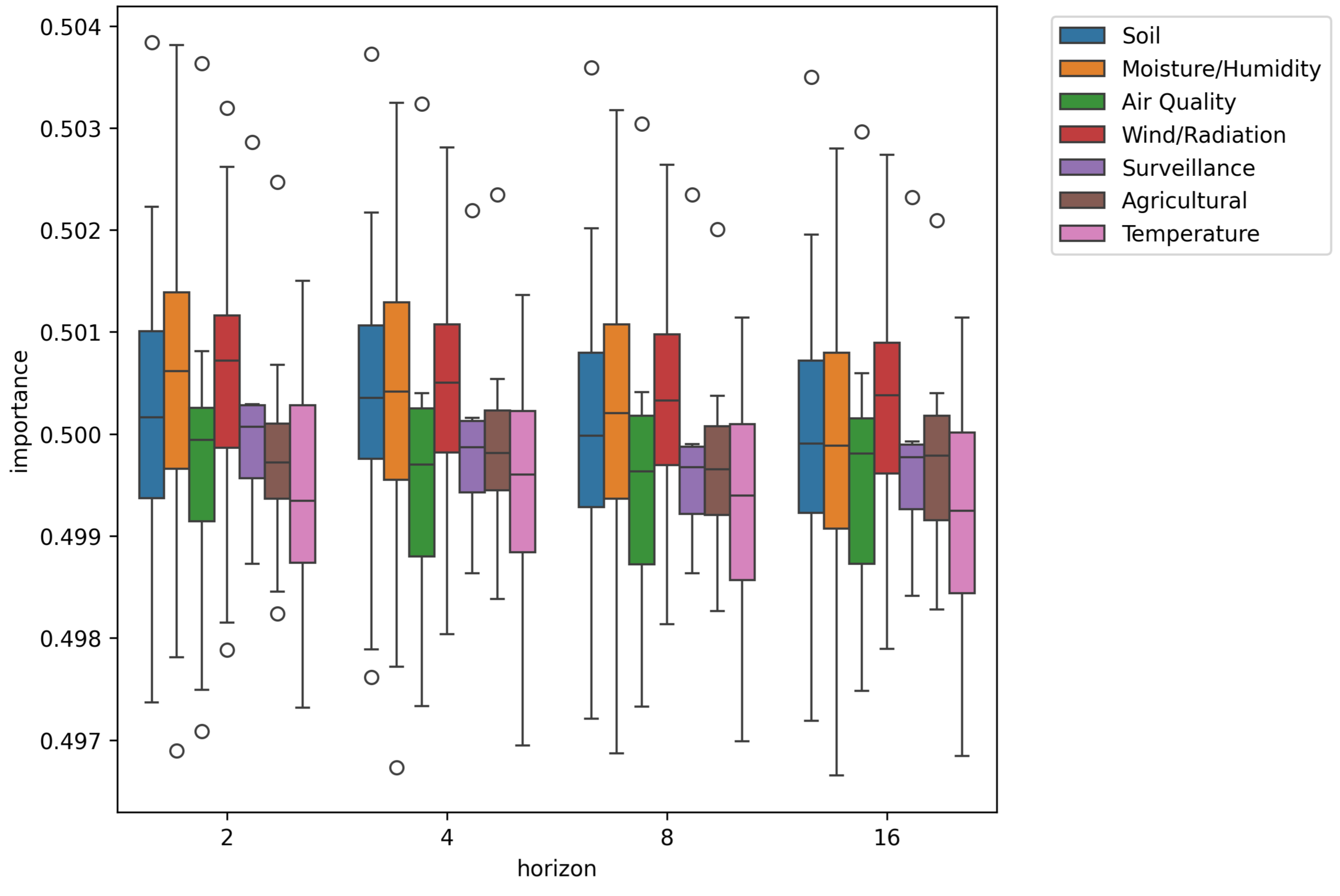}
\caption{Feature importance distribution by environmental category across all forecast horizons, based on results from 100 random seeds.}
\label{fig:feature_importance_by_category}
\end{figure}

Figure~\ref{fig:feature_importance_by_category} illustrates the distribution of feature importance scores across environmental categories for all forecast horizons. The box plots demonstrate that feature importance remains remarkably consistent across categories and horizons.

\subsection{Forecasting Results}\label{ssec:forecasting_results}

A forecasting model was trained on data from the \emph{in-sample} period ending one epidemiological week prior to the rolling test set and used to generate predictions of Valley Fever cases in Maricopa County at multiple horizons: 2, 4, 8, and 16 weeks ahead.
For each forecast horizon, the input layer received \emph{all} predictors that were selected in the top 10\% of permutation importance rankings across 100 random seeds (19 variables in total; see Table~\ref{tab:top_features_all}).
During evaluation, the input window was advanced weekly (stride = 1), while model weights remained \emph{fixed}; only the lagged covariates and target values changed.
This fixed-model walk-forward protocol ensures that each test observation remains strictly out-of-sample while avoiding the computational cost of retraining at each step.
Table~\ref{tab:performance_metrics_all} summarizes the forecasting performance across all evaluated horizons over the test period.

\begin{table}[h]
\centering
\footnotesize
\caption{Forecasting performance across multiple horizons, aggregated over the rolling test set (epi-weeks 900–991)}
\label{tab:performance_metrics_all}
\begin{tabular}{lcccc}
\hline
\textbf{Metric} & \textbf{2-week} & \textbf{4-week} & \textbf{8-week} & \textbf{16-week} \\
\hline
Mean Absolute Percentage Error (MAPE) & 0.13 & 0.16 & 0.21 & 0.23 \\
Mean Absolute Error (MAE)             & 24.35 & 29.77 & 37.36 & 42.71 \\
Mean Squared Error (MSE)              & 1\,018.83 & 1\,525.78 & 2\,432.47 & 3\,075.13 \\
Root Relative Squared Error (RSE)     & 7.55 & 2.25 & 1.73 & 1.47 \\
\hline
\end{tabular}
\end{table}

\begin{figure}[htb]
  \centering
  \includegraphics[width=\linewidth]{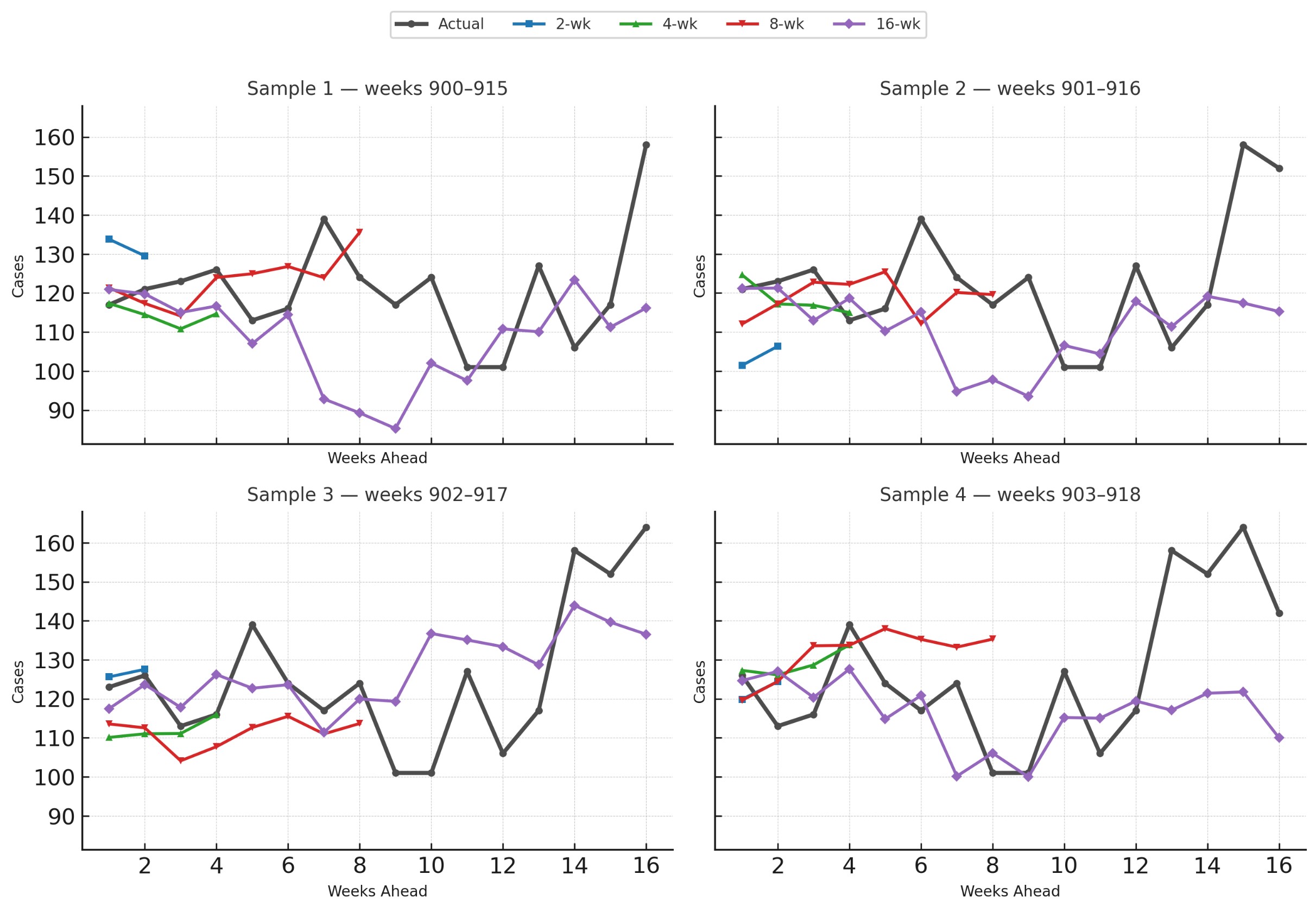}
  \caption{Rolling 16-week test samples (weeks 900 – 918). Black shows observed counts; colors show 2, 4, 8, and 16-week forecasts.
}
  \label{fig:rolling_forecasts}
\end{figure}

\paragraph{Visual inspection of rolling multi-horizon forecasts}
Figure~\ref{fig:rolling_forecasts} displays four consecutive 16-week test samples  
(Sample 1: epi-weeks 900–915; Sample 2: 901–916; Sample 3: 902–917;  
Sample 4: 903–918).  
The solid \textbf{black} line shows the observed surveillance case counts,  
while coloured lines with markers present the model’s forecasts at 2-, 4-, 8-,  
and 16-week lead times.

In Sample 1 (weeks 900–915), the model successfully predicts the sharp increase  
around week 906 and the decline that follows. Short-term forecasts closely track  
the observed pattern, while the 16-week forecast slightly underestimates the  
height of the rise.

Sample 2 (weeks 901–916) shows a gradual upward trend, which the model captures  
consistently across all horizons. Shorter forecasts match the observed levels  
more precisely, while longer horizons smooth over some of the local variation.

In Sample 3 (weeks 902–917), the model reflects the steady levels early in the  
window and correctly anticipates the rise after week 913. Short-term forecasts  
align closely with observed values, and even the 16-week forecast captures the  
directional change despite some underestimation.

Sample 4 (weeks 903–918) presents larger swings, with a decline in the middle  
and an increase at the end. While some overestimation appears at the 8-week  
horizon near the final peak, all forecast horizons capture the general sequence  
of changes.

Overall, these visual results reinforce the quantitative findings in  
Table~\ref{tab:performance_metrics_all}. The model not only maintains reasonable  
error levels across short- and mid-range horizons but also provides valuable  
early warning of rising or falling case trends up to four months in advance,  
supporting timely public health planning and response.

\section{Conclusion}
This paper presents a Graph Neural Network architecture for forecasting coccidioidomycosis (Valley Fever) incidence in Arizona, establishing the first successful implementation of graph-based deep learning for environmental disease prediction. The proposed hybrid architecture integrates graph-based feature selection with Transformer-based temporal modeling to capture complex spatiotemporal dependencies and lagged effects in disease transmission dynamics.

This architecture achieves robust forecasting performance across multiple prediction horizons, with Mean Absolute Percentage Errors ranging from 13\% for 2-week forecasts to 23\% for 16-week forecasts. This performance enables reliable early warning capabilities extending up to four months in advance, providing substantial lead time for public health planning and resource allocation. The consistency of performance across different forecast horizons demonstrates the model's effectiveness in capturing both short-term fluctuations and longer-term epidemiological trends.

This architecture eliminates the need for manual feature engineering by automatically identifying relevant environmental predictors and their temporal relationships. This automation reduces the expertise required for model deployment while ensuring consistent performance across different environmental conditions and geographic contexts.



\section*{Acknowledgments}

Beckett Sterner and Petar Jevti\'c were supported in part by National Institutes of Health, United States grant DMS-1615879.

\bibliographystyle{unsrtnat}
\bibliography{references}

\end{document}

%% file: Acro.tex
\begin{acronym}[paper]
\acro{GNN}{Graph Neural Network}
\acro{LSTM}{Long Short-Term Memory}
\acro{AR}{Auto-Regressive}
\acro{MA}{Moving Average}
\acro{VAR}{Vector Auto-Regression}
\acro{SARIMA}{Seasonal Auto-Regressive  Integrated  Moving  Aver-age}
\acro{VARMA}{Vector Auto-Regression Moving Average}
\acro{SES}{Simple Exponential Smoothing}
\acro{HWES}{Holt Winter’s Exponential Smoothing}
\acro{ML}{Machine Learning}
\acro{RF}{Random Forest}
\acro{GC-LSTM}{Graph Convolutional LSTM}
\acro{GCN}{Graph Convolutional Network}
\acro{GAT}{Graph Attention Network}
\acro{GGNN}{Gated Graph Neural Network}
\acro{MPNN}{Message Passing Neural Network}
\acro{RNN}{Recurrent Neural Network}
\acro{TCN}{Temporal Convolutional Network}
\acro{tsCV}{time series cross-validation}
\acro{RSE}{Root Relative Squared Error}
\acro{CORR}{Empirical Correlation Coefficient}
\acro{GAT-LSTM}{Graph Attention Long Short-Term Memory}
\acro{AutoML}{Automated Machine Learning }
\acro{EMD}{Empirical mode decomposition}
\acro{MSE}{Mean Squared Error}
\acro{MAE}{Mean Absolute Error}
\acro{GIN}{Graph Isomorphism Network}
\acro{OOS}{out-of-sample}

\end{acronym}